\documentclass[letterpaper, 10 pt, conference]{ieeeconf}  

\IEEEoverridecommandlockouts                              

\usepackage{graphics} 
\usepackage{epsfig} 
\usepackage{mathptmx} 
\usepackage{times} 
\usepackage{amsmath} 
\usepackage{amssymb}  

\title{\LARGE \bf
Tactile-based Reinforcement Learning for Adaptive Grasping under Observation Uncertainties
}

\author{Xiao Hu$^{1}$ , Yang Ye$^{1}$
\thanks{$^{1}$Xiao Hu is PhD Student of 
        Northeastern University, Massachusetts, Boston, USA,
        {\tt\small 
        xiao.h1@northeastern.edu}}%
\thanks{$^{1}$Gilbert Ye is Assistant Professor of
        Northeastern University, Massachusetts, Boston, USA,
        {\tt\small ye.y@northeastern.edu}}%
}

\begin{document}

\maketitle
\thispagestyle{empty}
\pagestyle{empty}

\begin{abstract}
Robotic manipulation in industrial scenarios such as construction commonly faces uncertain observations in which the state of the manipulating object may not be accurately captured due to occlusions and partial observables. For example, object status estimation during pipe assembly, rebar installation, and electrical installation can be impacted by observation errors. Traditional vision-based grasping methods often struggle to ensure robust stability and adaptability. To address this challenge, this paper proposes a tactile simulator that enables a tactile-based adaptive grasping method to enhance grasping robustness. This approach leverages tactile feedback combined with the Proximal Policy Optimization (PPO) reinforcement learning algorithm to dynamically adjust the grasping posture, allowing adaptation to varying grasping conditions under inaccurate object state estimations. Simulation results demonstrate that the proposed method effectively adapts grasping postures, thereby improving the success rate and stability of grasping tasks.

\end{abstract}

\section{INTRODUCTION}
Robotic manipulation in construction has witnessed rapid advancements driven by improvements in perception, control, and learning-based methods \cite{c1}. For example, researchers have been exploring generalizable grasping policies in construction objects \cite{c2}, dexterous manipulation in pipeline assembly \cite{c3}, and grasping deformable construction materials \cite{c4}. Traditional grasping approaches typically rely on observing the current scene, computing an optimal grasp pose, and then executing the grasp accordingly. However, these methods inherently depend on a strong assumption—that the robotic gripper can interact with the object precisely in the planned grasp pose, which can be far off in reality, especially in dynamic and uncertain construction settings. Specifically, most grasp generation models assume that the gripper can reach the desired pose perfectly and that the object pose is correctly captured and remains static during the grasping process. Such assumptions are often unrealistic in reality for the dynamic grasping process, especially during the closing motion of the gripper, in which interactions between the gripper and the object can lead to unintended perturbations. These factors significantly impact grasp stability and success rates in real-world scenarios.

To address these challenges, we propose a tactile-based adaptive grasping approach that enables the gripper to dynamically adjust its grasp in response to real-time tactile feedback. Specifically, we designed a tactile feedback simulator in MuJoCo \cite{c5} to facilitate efficient policy training. The simulator generates real-physics tactile sensor data based on force interaction and geometric features. Using Proximal Policy Optimization (PPO), we train a reinforcement learning policy that governs the adaptive behavior of the gripper. To bridge the gap between simulated and real-world tactile data, We introduce downsampled tactile observations, which refer to using uniformly spaced sampling as the input for the policy, ensuring robustness and generalization across different tactile sensing conditions. An experiment is performed to evaluate the design of the adaptive grasping policy based on tactile data. We simulate objects of different poses and postures, and add random noise to the pose observation input to replicate the uncertain observation in real manipulation tasks. By incorporating tactile feedback in this manner, our approach improves grasp stability and robustness, effectively mitigating the limitations of traditional grasp planning methods.

\section{Related Work}

\subsection{Grasp Planning Algorithms
}

Previous robotic grasping methods are mainly based on vision-based perception and analytical grasp planning \cite{c6,c7}. Early approaches focus on computing force closure and form closure grasps using geometric models of objects, assuming that accurate object shape and pose information is available \cite{c8,c9}. These methods typically use deterministic grasp planners that generate a fixed grasp pose, assuming the gripper can reach the computed pose without interference. However, real-world grasping tasks often involve uncertainties such as incomplete object perception, occlusions, and dynamic interactions, which reduce the effectiveness of purely model-based approaches \cite{c10,c11}.

To address these limitations, data-driven grasping techniques have gained popularity. Deep learning-based grasp planning methods leverage large-scale datasets to learn grasp success predictions \cite{c12,c13}. Grasping models such as GraspNet \cite{c14} and GQ-CNN \cite{c15} have demonstrated promising performance in generating robust grasps from RGB or depth images. However, these methods still assume that the gripper can execute the planned grasp pose precisely and that the object remains static during the grasp execution, which is often unrealistic in practical scenarios.

\subsection{Reinforcement Learning for Grasping}

Reinforcement learning has been widely used in robotic manipulation tasks, including grasping, regrasping, and hand manipulation. Model-free RL algorithms such as Deep Q-Networks (DQN)\cite{c16} , and Proximal Policy Optimization (PPO) have been applied to learn grasping strategies from trial-and-error interactions. Recent works have explored RL-based closed loop grasping, where policies continuously refine grasp poses during execution . However, most existing methods rely only on visual feedback, which may not be reliable in scenarios with occlusions or noisy depth data.

Our work differs from previous RL-based grasping methods by focusing on tactile-driven policy learning. Instead of relying on full-state object observations, we use downsampled tactile signals as input, making the policy more robust to real-world sensor noise. Furthermore, our PPO-trained policy generates dynamic grasp adjustment parameters, allowing the gripper to refine its grasp adaptively throughout execution. This approach enables successful grasping even when initial grasp predictions are suboptimal, bridging the gap between static grasp planning and real-world grasp execution.

\section{Method}

This section primarily introduces the tactile feedback simulation framework and the use of tactile feedback for adaptive grasping policy training. The proposed framework enables high-fidelity physics-based tactile interaction modeling, facilitating realistic contact perception within a simulated environment. Furthermore, we explore how tactile information can be effectively integrated into policy learning, allowing for dynamic and adaptive grasping strategies based on real-time sensory feedback.

\begin{figure}[htbp]
    \centering
    \includegraphics[width=0.9\linewidth]{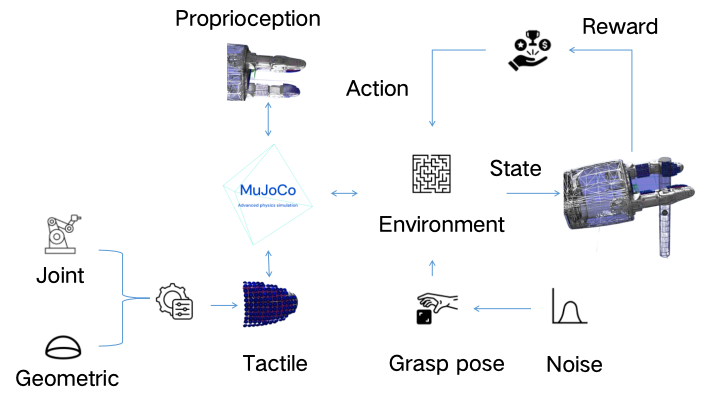}
    \caption{Framework}
    \label{fig:enter-label}
\end{figure}

As shown in Figure 1, we built a tactile simulation framework in MuJoCo, which includes a tactile simulation system and a robotic hand system. The tactile simulation encodes physical interactions such as friction, collision,  and inertia. The simulated tactile sensor arrays are placed on the surface of the robotic hand based on real-life robotic hand designs. In this study, we adopt the Proximal Policy Optimization (PPO) algorithm for reinforcement learning (RL) training. The observation space covers robotic preceptive data, initial (uncertain) observation of the object pose, and optional tactile feedback.

\subsection{Tactile Simulation Framework} 

To enable high-fidelity tactile interaction simulation, we develop a physics-based tactile sensing framework using MuJoCo. This framework integrates a rigid-body structure with MuJoCo’s sensor module, enabling the simulation of contact-based tactile perception. Specifically, when a designated site within the model makes contact with another object, MuJoCo computes the interaction forces at the contact point. A touch sensor is then employed to measure and output these forces as a scalar value, representing the magnitude of the tactile interaction.

\begin{figure}[htbp]
    \centering
    \includegraphics[width=0.45\linewidth]{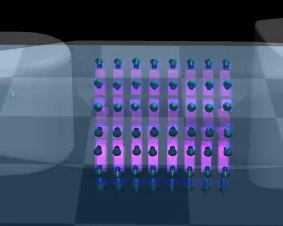}
    \caption{Tactile simulation framework}
    \label{fig:tactile-sim}
\end{figure}

The tactile sensor model is designed to ensure realistic contact interactions by incorporating a joint mechanism, a geometric structure, and dynamically stable inertial properties. The key components of the design are as follows:

\subsection{Joint Configuration for Tactile Deformation}
To simulate skin-like deformation upon contact, the tactile sensor body is equipped with a sliding joint (`slide`), allowing controlled displacement along the Z-axis. The primary joint parameters are defined as follows:

\begin{itemize}
    \item \textbf{Damping Coefficient:} Regulates energy dissipation, ensuring smooth movement in response to external forces.
    \item \textbf{Stiffness Parameter:} Determines the elastic response of the joint, allowing the sensor to return to its original position after contact.
\end{itemize}

\begin{align}
a_1 + d \cdot (b v + k r) = (1 - d) \cdot a_0
\end{align}

\(a_1\) denote the acceleration, \(v\) the velocity, \(r\) the position or residual (defined as \(O\) in friction dimensions), \(k\) and \(b\) the stiffness and damping of the virtual spring used to define the reference acceleration \(a_{\text{ref}} = -b v - k r\). Let \(d\) be the constraint impedance, and \(a_0\) the acceleration in the absence of constraint force. 

The formula illustrates how damping and stiffness affect acceleration, which in turn influences force and ultimately impacts tactile simulation.

\begin{itemize}
    \item \textbf{Rotational Inertia:} Affects the dynamic response of the sensor, influencing its reaction to applied forces.\\
    The moment of inertia is a symmetric \(3 \times 3\) matrix (unit: kg·m²), $\alpha$  is a parameter aimed at reducing the discrepancy between sim2real.representing the distribution of inertia around the three coordinate axes:

    \begin{align}
    \mathbf{I} = \alpha \circ
    \begin{bmatrix}
    I_{xx} & I_{xy} & I_{xz} \\
    I_{xy} & I_{yy} & I_{yz} \\
    I_{xz} & I_{yz} & I_{zz}
    \end{bmatrix}
    \end{align}
    \item \textbf{Motion Constraints:} Limits the joint’s displacement range to prevent excessive deformation while maintaining tactile sensitivity.

\end{itemize}

\subsection{Geometric Representation of the Tactile Sensor}
To ensure accurate force distribution and realistic contact dynamics, the tactile sensor is modeled using a mesh-based geometric structure with the following attributes:

\begin{itemize}
    \item \textbf{Contact Affinity Settings:} Defines interaction properties, specifying which objects can establish contact with the sensor.

    \begin{align*}
    &(\text{A.contype} \ \& \ \text{B.conaffinity}) \neq 0 \\
    &\quad \text{OR} \\
    &(\text{B.contype} \ \& \ \text{A.conaffinity}) \neq 0
    \end{align*}
        
    \item \textbf{Surface Friction Properties:} Determines resistance to tangential forces, influencing how the sensor interacts with surfaces of varying textures. ${f}_{\text{slide}}$ denotes the frictional force during sliding, $\mu_1$ is the friction coefficient, and $v_t$ represents the relative tangential velocity.
    \begin{align}
    \mathbf{f}_{\text{slide}} = -\mu_1 F_n \frac{\mathbf{v}_t}{|\mathbf{v}_t|}
    \end{align}

\end{itemize}

\subsection{Policy Training}
Specifically, we use tactile feedback along with the current proprioceptive state—including the joint angles of the robotic hand and the Cartesian pose of its end-effector—as input. The model then outputs the motion increments for the next time step, including joint angle variations and end-effector pose increments.

Furthermore, during the training process, we introduce random noise to the object’s position to enhance the robustness of the policy. More precisely, instead of placing the object exactly at a predefined grasping pose (including its coordinates and joint angles), we add noise to its position in the simulation environment. This perturbation helps improve the adaptability of the policy to environmental variations. To further enhance the generalization capability of the policy, we include objects of various common shapes, such as cuboids and spheres, ensuring that the trained policy can be applied to a broader range of tasks.

To enable stable and effective grasping behavior, we design a reward function that integrates tactile feedback, joint motion efficiency, and object pose consistency. The reward function is formulated to encourage \textit{stable contact with the object}, \textit{minimized unnecessary hand and finger movements}, and \textit{precise object manipulation}. It consists of the following key components.

\begin{figure}[htbp]
    \centering
    \includegraphics[width=0.3\linewidth]{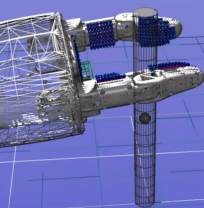}
    \caption{Adaptive grasp}
    \label{fig:enter-label}
\end{figure}

The overall reward function $R_{grasp}$ is then defined as:

\begin{align}
R_{grasp} &= Q_{fingertip}  
                       - Z_{hand} \\
                     & - Z_{fjoint}  
                       - O_{diff} \notag \\
                     &\quad - D_{diff}.
\end{align}

A \textit{contact reward term} is introduced to maintain stability during contact. \textit{Contact reward term} counts the number of fingertips that actively 
 ($T$) make contact with the object and poses a negative reward of amplitude \(\beta\) to encourage constant contact:

\begin{align}
Q_{fingertip} &= \beta \times T
\end{align}

\begin{align}
\begin{bmatrix}
Z_{fjoint} \\
Z_{hand} 
\end{bmatrix}
=
\begin{bmatrix}
|G_{diff}[7:]| \\
|G_{diff}[:3]| \times \gamma
\end{bmatrix}
\end{align}

Here, $G_{diff}$ represents the grasp configuration, encompassing both joint and pose differences relative to the target $Z_{fjoint}$ represents the \textit{joint deviation} of the fingers, while $Z_{hand} $ quantifies \textit{hand displacement} during grasping $\gamma$ is a scaling factor. Both terms act as penalties, ensuring that minimal movement is performed to maintain a stable grasp.

$D_{diff}$ represents the \textit{positional deviation} of the object from its target location, while $O_{diff}$ represents its \textit{orientation error}. These terms ensure that the object remains as close as possible to the intended grasping pose.This formulation encourages stable grasping by maximizing fingertip contact while simultaneously minimizing unnecessary movement and ensuring accurate object positioning. The function is periodically logged for evaluation, providing insights into the grasping process.

This structured reward function enables the \textit{learning of an adaptive grasping policy} by reinforcing stable and controlled interactions, ensuring efficient manipulation and tactile-based policy optimization.

\section{EXPERIMENT}
\subsection{Training}

\begin{figure}[htbp]
    \centering
    \includegraphics[width=0.7\linewidth]{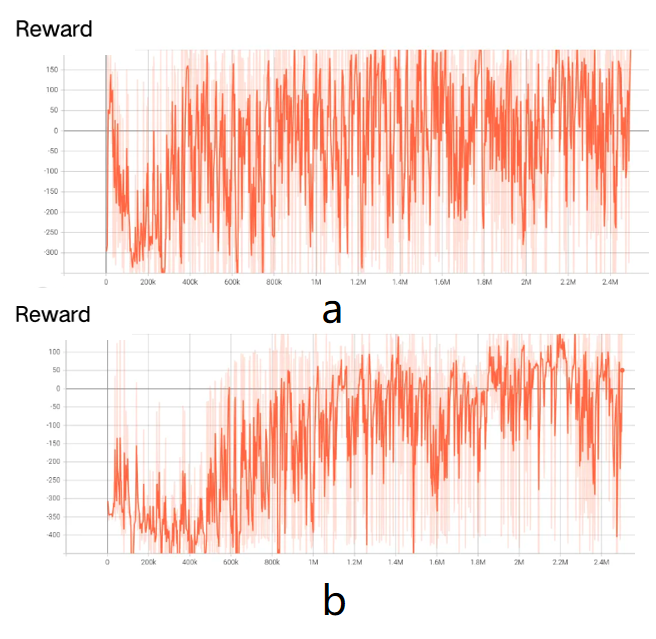}
    \caption{Training porcess: (a) represents the policy incorporating tactile feedback, whereas(b) depicts the policy trained without tactile.}
    \label{fig:training_curve}
\end{figure}

\begin{itemize}
  \item \textbf{Exploration Oscillation Phase} : In (a), the reward values exhibit significant fluctuations between $150$ and $-300$ before reaching 200k steps, indicating high-entropy exploratory behavior of the policy network. In contrast, (b) reflects a lower-entropy exploration compared to (a)
  
  \item \textbf{Policy Optimization Phase} : A distinct inflection point between 200k and 400k steps in (a) initiates a monotonic improvement. In contrast, (b) exhibits a more gradual improvement, beginning at 200k steps and extending to nearly 1.2M steps.
  
  \item \textbf{Convergence Plateau Phase} : the policy in (a) achieves stable performance, with reward values consistently maintained within the range of $[100, 150]$, while (b) stabilizes within $[0, 50]$. Both satisfy the convergence criterion:

  \begin{equation}
    \max_{ \Delta \geq 1k} \left| \frac{R_{t+i} - R_t}{R_t} \right| < 0.05
  \end{equation}
\end{itemize}

\subsection{Comparative experiment}

\begin{figure}[htbp]
    \centering
    \includegraphics[width=0.65\linewidth]{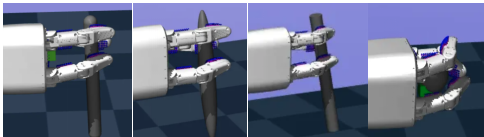}
    \caption{Grasping distinct objects}
    \label{fig:Grasping distinct objects}
\end{figure}

As shown in Figure~\ref{fig:Grasping distinct objects}, to systematically evaluate the efficacy of tactile sensing integration, we conducted controlled comparative experiments under two distinct experimental conditions: tactile-enabled (TE) and tactile-disabled (TD) configurations. The experimental protocol involved 100 independent trials per condition across four geometrically distinct objects (column, capsule, ellipsoid, and sphere). Crucially, these test objects were carefully designed with matched physical parameters (mass: ± 0.01 kg, volume: ± 5 cm³) to isolate shape variation as the primary generalization challenge.The experimental results demonstrate that incorporating tactile feedback significantly improves the success rate, further confirming the importance of tactile perception in this task. Therefore, integrating tactile information is both meaningful and necessary for enhancing system performance.

\begin{table}[h]
    \centering
    \begin{tabular}{c|c|c}
         & Tactile & No tactile \\ 
        \hline
        Column & 95\% & 87\% \\ 
        \hline
        Capsule & 96\% & 88\% \\ 
        \hline
        Ellipsoid & 92\% & 89\% \\ 
        \hline
        Sphere & 93\% & 82\% \\ 
    \end{tabular}
    \caption{experiment}
    \label{tab:example}
\end{table}

\section{CONCLUSIONS}
 
Experimental results demonstrate that our method enables the gripper to dynamically adapt its grasp pose in response to object perturbations, significantly improving grasp success rates in both simulation and real-world applications.

In summary, the proposed tactile-based adaptive grasping framework effectively addresses the limitations inherent in traditional grasping approaches. By leveraging tactile feedback, our method enables robust grasping even in scenarios where visual occlusions occur, mitigating the reliance on purely vision-based perception. Furthermore, the framework enhances grasp stability by dynamically adjusting to unexpected object perturbations, ensuring reliable manipulation in unstructured environments. These results highlight the potential of integrating tactile sensing into robotic grasping strategies, paving the way for more adaptive and resilient manipulation systems.

\addtolength{\textheight}{-12cm}   




\section*{ACKNOWLEDGMENT}

We sincerely appreciate Tencent Robotics X Lab and extend our gratitude to Bidan Huang for their contributions.

\end{document}